\documentclass[sigconf]{aamas} 

\usepackage{balance} % for balancing columns on the final page
\usepackage{graphicx}
\usepackage{subcaption}
\usepackage{algorithm}
\usepackage{algpseudocode}

\usepackage{xcolor}
\definecolor{qquestion}{rgb}{0.02, 0.75, 0.2}
\definecolor{ssuggestion}{rgb}{0.0, 0.0, 1.0}
\definecolor{ffix}{rgb}{1.0, 0.2, 0.2}
\definecolor{rreview}{rgb}{0.9, 0.63, 0}
\definecolor{llongterm}{RGB}{74, 80, 133}
\definecolor{myred}{rgb}{1.0, 0.2, 0.2}
\definecolor{todo}{rgb}{0.98,0.925,0.764}
\definecolor{lightbrown}{rgb}{0.85, 0.25, 0.25}

\newcommand{\eat}[1]{} %comment out paragraphs

\setcopyright{none}
\acmConference[ALA '26]%
{Proc.\@ of the Adaptive and Learning Agents Workshop (ALA 2026)}%
{May 25 -- 26, 2026}%
{Paphos, Cyprus, https://alaworkshop2026.github.io/}%
{Aydeniz, Delgrange, Mohammedalamen, Yang (eds.)}%
\copyrightyear{2026}
\acmYear{2026}
\acmDOI{}
\acmPrice{}
\acmISBN{}
\acmSubmissionID{58}

\title[S3]{S3: Stable Subgoal Selection by Constraining Uncertainty of Coarse Dynamics in Hierarchical Reinforcement Learning}

\author{Kshitij Kumar Srivastava}
\affiliation{
  \institution{University of Massachusetts, Lowell}
  \country{United States}}
\email{kshitijkumar_srivastava@uml.edu}

\author{Kshitij Jerath}
\affiliation{
  \institution{University of Massachusetts, Lowell}
  \country{United States}}
\email{kshitij_jerath@uml.edu}

\begin{abstract}
Hierarchical Reinforcement Learning (HRL) intends to separate strategic planning from primitive execution. It has been widely successful in solving long-horizon and complex tasks, where flat-RL algorithms have difficulty in learning. However, while the low-level agent in HRL benefits from dense feedback and abundant trial opportunities, the high-level agent receives sparse, delayed feedback from the environment and its performance depends on the low-level execution capability. In this paper, we study whether subgoal selection by the high-level agent can be performed 
more strategically, by providing it with dynamics-aware intrinsic motivation.
Since motivation based on primitive transition dynamics would require broad coverage of the state-action space, we propose to use \emph{coarse dynamics}, i.e., environment transitions aggregated over multiple steps at the temporal scale at which the high-level agent operates.
This approach stabilizes the high-level policy by learning to minimize the predictive uncertainty associated with the coarse dynamics, and provides a guided structure for navigation.
We model the predictive uncertainty by evaluating different dispersion metrics as approximated by a Mixture Density Network (MDN). Empirically, we observe that a dense, dynamics-aware intrinsic reward leads to risk-averse subgoal selection, enabling it to outperform state-of-the-art HRL methods in non-stationary long-horizon environments.
\end{abstract}

\newcommand{\BibTeX}{\rm B\kern-.05em{\sc i\kern-.025em b}\kern-.08em\TeX}

\begin{document}

%%% The following commands remove the headers in your paper. For final 
%%% papers, these will be inserted during the pagination process.

\pagestyle{fancy}
\fancyhead{}

%%% The next command prints the information defined in the preamble.

\maketitle 

%%%%%%%%%%%%%%%%%%%%%%%%%%%%%%%%%%%%%%%%%%%%%%%%%%%%%%%%%%%%%%%%%%%%%%%%

\section{Introduction}

\begin{figure*}[!tbp]
  \centering
  \begin{subfigure}[t]{.70\textwidth}
    \centering
    \includegraphics[width=\linewidth]{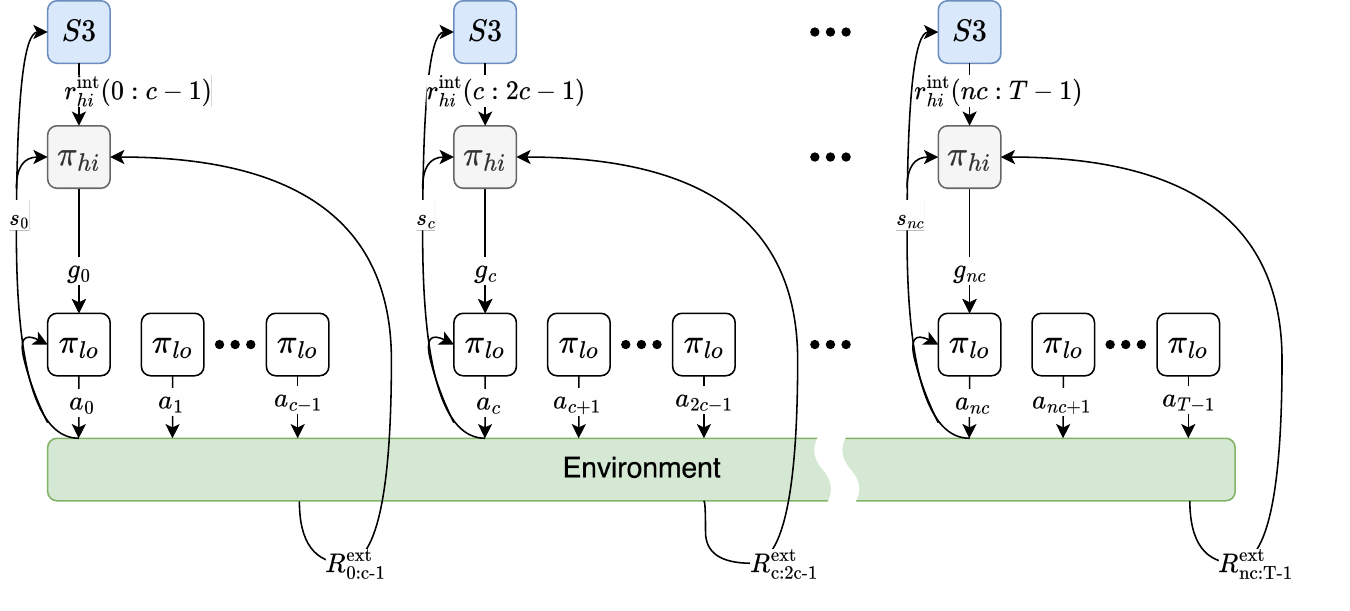}
    \subcaption{}\label{fig:s3-a}
  \end{subfigure}\hfill
  \begin{subfigure}[t]{.25\textwidth}
    \centering
    \includegraphics[width=\linewidth]{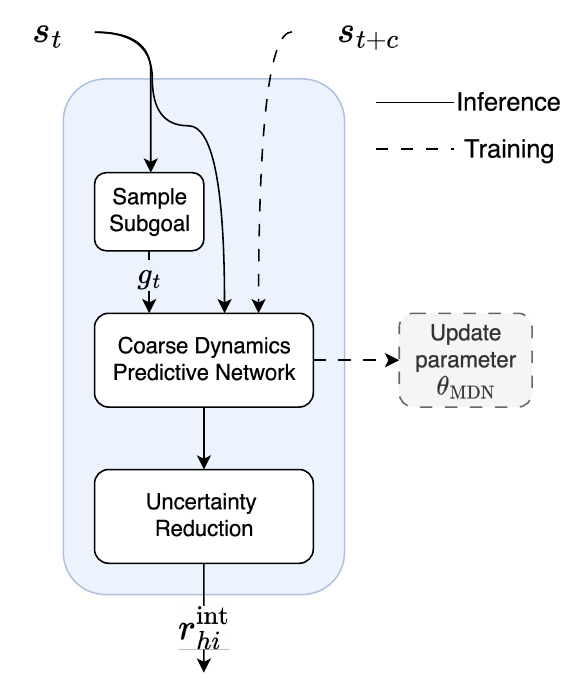}
    \subcaption{}\label{fig:s3-b}
  \end{subfigure}
  \caption{HRL framework with S3 implemented by using MDN Network. (b) S3 module implementation. During the training, S3 updates the MDN parameters $\theta_{\text{MDN}}$ based on supervised learning using the tuple $(s_t, g_t, s_{t+c})$. It samples $g_t$ from $\pi_{\text{hi}}$ and $s_t$ from environment during the execution phase.}
  \label{fig:s3}
\end{figure*}
Hierarchical Reinforcement Learning (HRL) breaks down complex tasks into multiple subtasks, improving the tractability of solving long-horizon tasks, with the capability of handling sparse environmental rewards \cite{barto2003recent, dayan1992feudal}. It leverages a hierarchical framework that enables one agent (low-level agent or `Worker') to learn the fine-grained details of completing a simple task, while another agent (high-level agent or `Manager') concurrently learns the coarse-grained objective of assigning these tasks or subgoals at some predefined intervals.
While the efficacy of HRL in achieving a specific goal hinges on a well-calibrated hierarchical interface, concurrent training creates non-stationarity for both the low-level agent, which has to contend with changing high-level subgoal assignments that it needs to learn to achieve, and the high-level agent, which needs to learn how to assign subgoals in the face of the changing ability of the low-level agent to complete those subgoals.

Recent advances in hierarchical approaches \cite{nachum2018data, levy2017learning} have sought to overcome the challenge of non-stationarity by employing Hindsight Experience Replay (HER) \cite{andrychowicz2017hindsight}. This approach helps the high-level policy to learn by assuming that the behavior of the low-level policy (even if unsuccessful) corresponds to the actual optimal policy. In this way, the learning of the high-level agent is effectively guided by a low-level policy that \textit{appears} to be stationary. Other work has built upon this in various ways, such as constraining the high-level agent actions (i.e., subgoal assignments) to states that are \textit{reachable} by the low-level agent \cite{zhang2020generating}, or focusing on specific landmarks that are \textit{promising} and \textit{novel} and may assist in learning the optimal policy \cite{kim2021landmark}.

However, we believe that these approaches address only the symptoms of non-stationarity -- they mask a deeper insight that is fundamental to the relationship and learning objectives of the Manager-Worker team. Specifically, we argue that to enhance the ability of the Manager to learn, the subgoal assignment approach should value subgoals that are not just reachable, but also \textit{predictive}, i.e., those that the low-level agent is expected to reach with reasonable certainty -- an important consideration for promoting stationarity. In other words, the high-level agent should motivate its policy by explicitly learning a coarse predictive model of the environment, by means of learning the predictable transition dynamics of the low-level agent.
We introduce the notion of Stable Subgoal Selection (S3), wherein the coarse predictive dynamics indicate a strategic-level model that abstracts away fine-grained primitives and retains only the signal relevant to high-level decisions regarding subgoal selection (Section \ref{Sec:Theoretical Analysis}). Thus, S3's intrinsic reward is conditioned on the competence or predictability of the low-level policy, since the subgoals assigned by the high-level agent are only useful if the low-level agent can execute them reliably.
For example, in a navigation task, if the Manager is uncertain about the Worker squeezing through a narrow doorway but confident that it would reach a nearby hallway junction, it would prefer the low-uncertainty junction limits the planning error and yields predictably executable subgoals. We would like to clarify that both subgoals (the narrow doorway and hallway junction) are reachable, but one is more likely to be reached than the other, i.e., there is sufficient predictive certainty associated the hallway junction -- an indicator that this subgoal would successfully model the coarse dynamics of the system.

From an implementation perspective, we approximate coarse dynamics with the help of a Mixture Density Network (MDN) (Section \ref{Sec:MDN_Intrinsic_Reward}), which is trained in parallel with the high-level and low-level agent. We have used MuJoCo environments \cite{todorov2012physics} to benchmark and analyze the efficacy of S3. Empirical results exhibit S3 outperforms state-of-the-art HRL methods in environments with high uncertainty and long-term dependencies (Section \ref{Sec:Empirical Study}). Our approach is generalizable to HRL algorithms since it primarily focuses on the high-level intrinsic reward. In this work, we demonstrate the utility of incorporating our coarse predictive certainty framework into an existing HRL algorithm such as Hierarchical Reinforcement learning with k-step Adjacency Constraint (HRAC) \cite{zhang2020generating} that itself uses reachability-based subgoal assignment as a motivating principle.

%%%%%%%%%%%%%%%%%%%%%%%%%%%%%%%%%%%%%%%%%%%%%%%%%%%%%%%%%%%%%%%%%%%%%%%%
\section{Preliminaries}
\label{Sec:Preliminaries}

We model this problem as a goal-conditioned Markov Decision Process (MDP), 
$\mathcal M=( S,\;A,\;P,\;R^{\mathrm{ext}},\;\gamma,\;\mathcal G_{\text{task}})$. 
Here, $S\subseteq \mathbb{R}^{d_s}$ denotes the continuous state space, and $A\subseteq\mathbb{R}^{d_a}$ the continuous action space. The environment behaves according to the transition kernel $P(s'|s,a) = Pr(S_{t+1} = s' | S_t = s, A_t = a)$, and the agent seeks to maximize cumulative discounted reward with discount factor $\gamma\in(0,1)$.
At the start of each episode, a \emph{task goal} $g_{\text{task}}\sim\mathcal{G}_{\text{task}}$ is sampled from a goal distribution and concatenated to the state observation. The extrinsic reward function $R^{\text{ext}}(s, a, s')$ is typically sparse and provides a positive signal only when the agent reaches the goal. 

A sparse feedback limits classical reinforcement learning ability particularly in continuous control tasks where exploration in expansive state-action space is ineffective. To help mitigate this, the control of the MDP is extended hierarchically by introducing two distinct temporal levels of decision-making (Figure \ref{fig:s3-a}). 
The high-level agent learns a policy $\pi_{\text{hi}}(g\mid s; \theta_{hi})$ to assign subgoal $g_t$ every $c$ steps to the low-level agent, where $\theta_{\text{hi}}$ represents the parameters of the learned high-level policy.
The low-level agent operates at the environment's native time scale, it observes state $s_t$ and subgoal assigned by the manager $g_t$ to executes for the next $c$ steps, reaching the 'landing state' $s_{t+c}$. The Manager outputs a relative subgoal $g_t$ in state space and the Worker is trained to reach the target state $s_t + g_t$. The policy $\pi_{\text{lo}}(a\mid s, g ; \theta_{\text{lo}})$ is optimized with where $\theta_{\text{lo}}$ as the parameters of the learned low-level policy. The reward for the worker is a function of how `close' it has reached to the assigned goal i.e., $r^\text{lo}_t = -|| s_t -g_t +s_{t+c}||_2$, whereas the manager is rewarded by the environment and our intrinsic reward term. Formally, high-level decision times are $t\in\{0, c, 2c, \dots\}$. This separation of timescale allows the manager to focus on abstract, long-horizon planning while delegating fine-grained control to the worker.
%%%%%%%%%%%%%%%%%%%%%%%%%%%%%%%%%%%%%%%%%%%%%%%%%%%%%%%%%%%%%%%%%%%%%%%%
\section{Related Work}
\label{Sec:Related work}
Hierarchical Reinforcement Learning (HRL) and intrinsic motivation have been extensively explored as means to address sparse rewards, exploration inefficiency, and credit assignment in long-horizon tasks. In this section, we discuss the distinguishing features of our approach in the context of three main research directions: (i) hierarchical control and subgoal discovery, (ii) intrinsic reward shaping and potential-based methods, and (iii) information–theoretic formulations of skill learning.

Classical HRL frameworks such as options and feudal learning \cite{sutton1999between, dayan1992feudal} introduced temporal abstraction by allowing high-level controllers to invoke temporally extended actions. More recent methods \cite{nachum2018data, levy2017learning} operationalize this idea through a two-level hierarchy, where the higher-level policy issues subgoals expressed in the state-space, and the low-level policy executes them through continuous actions. These methods demonstrate substantial gains in sample efficiency and exploration over flat RL, yet they often assume a deterministic mapping between a current state, subgoal and the resulting terminal state achieved by the lower level.
For example, HIerarchical Reinforcement learning with Off-policy correction(HIRO) \cite{nachum2018data} introduces the idea of off-policy correction; the unsuccessful low-level trajectories (i.e., those that did not reach the subgoal $g_t$) are stored in a replay buffer. They are then relabeled so that reached state $s_{t+c} (\neq g_t)$ is given the label $g'_t$, indicating that this was the subgoal that was intended to reach. This approach makes the low-level agent policy appear to be stationary and makes the learning process more sample efficient.
Similarly, HRAC \cite{zhang2020generating} limits the high-level agents action space to a $k$-step adjacency region. It eliminates infeasible subgoals while preserving the optimality. 
HIerarchical reinforcement learning Guided by Landmarks(HIGL) \cite{kim2021landmark} focuses on reducing high-level action space by generating landmarks that optimize for broad coverage and novelty. Broadly, they also focus on modulating Manager's behavior.
\cite{nachum2018data, levy2017learning, nachum2019multi, gao2024hierarchical, zhang2020generating}. 
On the contrary, relatively few works have provided intrinsic rewards directly to the high-level policy, though recent work by Wang et al. \cite{wang2025hierarchical} has sought to address this issue by modulating the high-level objective via epistemic uncertainty to encourage exploration. They treat uncertainty as a signal of what the agent has not yet explored, guiding the Manager toward unfamiliar regions of the state space using a diffusion model.
In contrast, S3 treats this mapping as stochastic and explicitly models the distribution of possible outcomes under a given subgoal. This allows us to measure how consistently the Worker can realize the Manager's intent and shape high-level rewards using Stable Subgoal Selection. By incorporating uncertainty, using reward shaping directly through the conditional subgoal distribution, our method provides stability in spite of evolving low-level policy.

\emph{Reward shaping} is a classical technique for accelerating learning without altering the optimal policy, provided the shaping function satisfies the potential-based condition \cite{ng1999policy}. This principle has been used to guide exploration, to inject domain knowledge, or to improve credit assignment in hierarchical systems \cite{brys2015policy, devlin2014potential}. 
Reward-Sharing Relational Networks (RSRN) \cite{haeri2022reward} introduce a directed relational graph that parameterizes how much each agent “cares about” other agents’ outcomes and uses these relation weights to transform environment rewards into shared relational rewards. The resulting reward-sharing induces coordinated behaviors by coupling agents’ objectives through the learned or specified relations. We use this lens to design an intrinsic reward for the high-level agent that promotes reliable manager–worker coordination via predictable coarse outcomes.  
In hierarchical contexts, most shaping approaches focus on the low-level policy; defining intrinsic rewards as the negative distance between the current state and the subgoal \cite{nachum2018data, vezhnevets2017feudal, levy2017learning, watanabe2022shiro}. However, the high-level policy is typically left unshaped, relying solely on sparse extrinsic signals. Our work extends potential-based shaping to the manager’s level by deriving an intrinsic reward proportional to the variance of the goal-conditioned coarse dynamics. The resulting term remains policy-invariant but supplies dense, capability-aware feedback that improves exploration efficiency and training stability.

\emph{Information–theoretic frameworks} for unsupervised skill discovery, such as DIAYN \cite{eysenbach2018diayn}, empowerment \cite{klyubin2005empowerment}, and VIC \cite{gregor2016variational}, aim to maximize mutual information (MI) between latent skills and resulting states. These methods encourage the discovery of diverse, distinguishable behaviors by maximizing mutual information between the landing state and subgoal. Optimizing mutual information can be expressed as maximizing the entropy of landing state, i.e., pushing for diverse outcomes and minimizing the conditioned entropy of the landing state given subgoal $I(G; S_{t+c}) = H(S_{t+c}) - H(S_{t+c} | G)$), i.e., having more certain outcomes for a subgoal.
Our method focuses specifically on the conditional term $H(S_{t+c}|G)$, which captures the reliability or predictability of the outcome given a chosen subgoal.
By penalizing the total variance of the coarse dynamics distribution, we effectively minimize an upper bound on this conditional entropy, thereby increasing the mutual information between the Manager’s subgoal and the achieved terminal state. Unlike MI-based methods, however, our formulation is computationally tractable, requiring only a single scalar moment (variance) from a learned mixture-density model, and maintains optimal-policy invariance through its potential-based structure.

%%%%%%%%%%%%%%%%%%%%%%%%%%%%%%%%%%%%%%%%%%%%%%%%%%%%%%%%%%%%%%%%%%%%%%%%
\section{Theoretical Analysis}
\label{Sec:Theoretical Analysis}

\begin{figure*}[t]
  \centering
  \begin{subfigure}[t]{.33\textwidth}
    \centering
    \includegraphics[width=\linewidth]{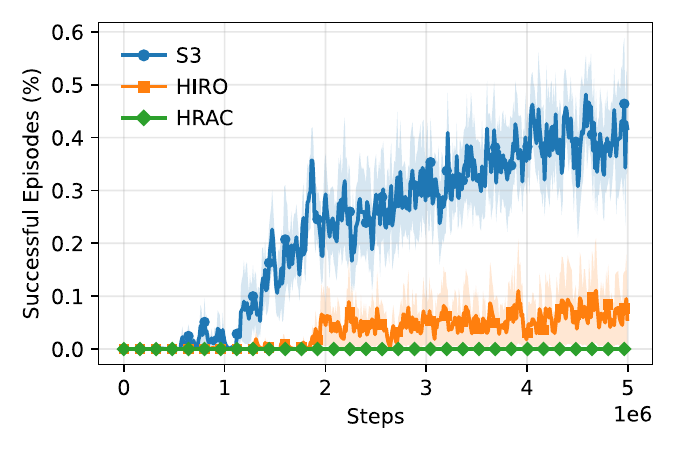}
    \includegraphics[width=0.9\linewidth]{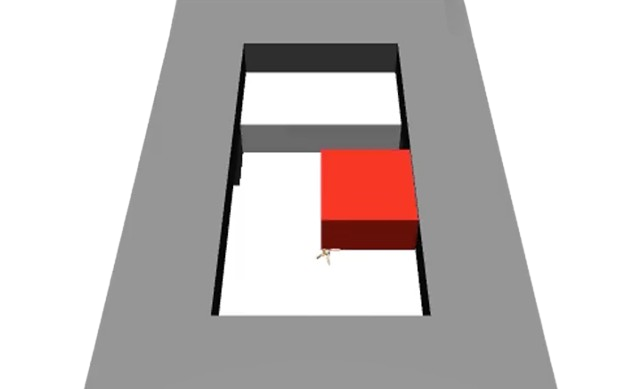}
    \subcaption{ \: Ant Fall}\label{fig:ant-a}
  \end{subfigure}\hfill
  \begin{subfigure}[t]{.33\textwidth}
    \centering
    \includegraphics[width=\linewidth]{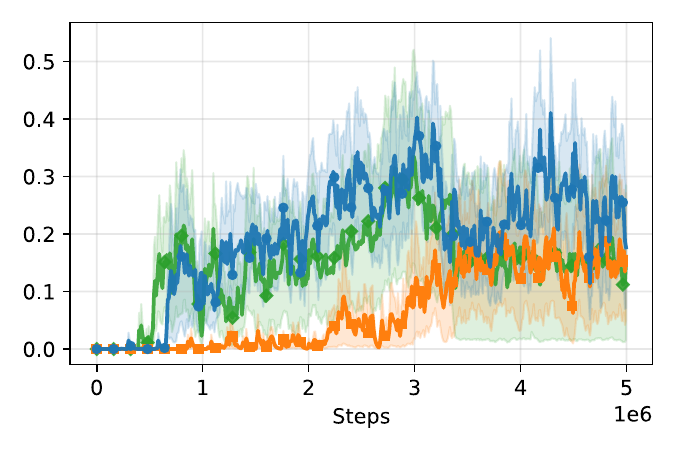}
    \includegraphics[width=0.9\linewidth]{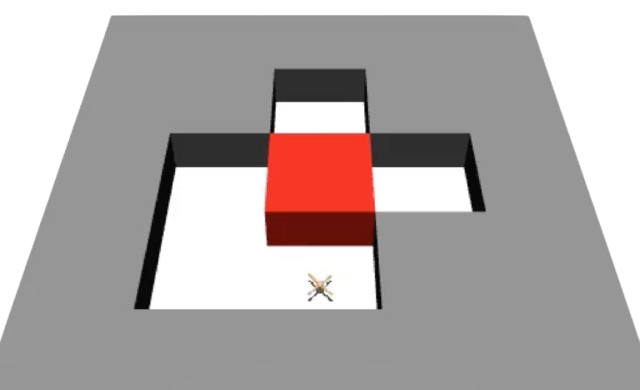}
    \subcaption{ \:Ant Push}\label{fig:ant-b}
  \end{subfigure}
  \begin{subfigure}[t]{.33\textwidth}
    \centering
    \includegraphics[width=\linewidth]{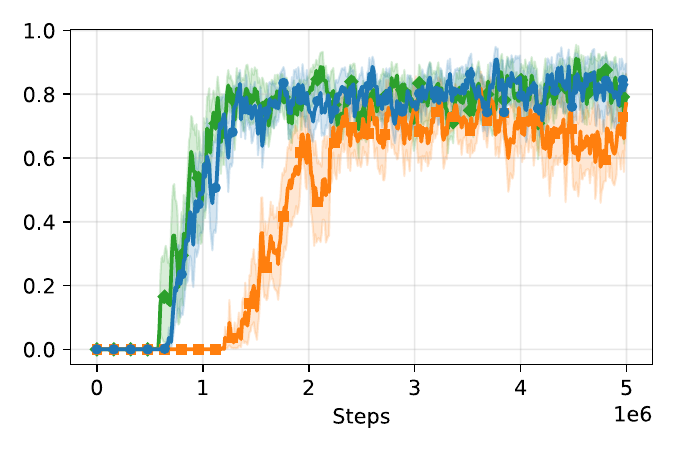}
    \includegraphics[width=0.9\linewidth]{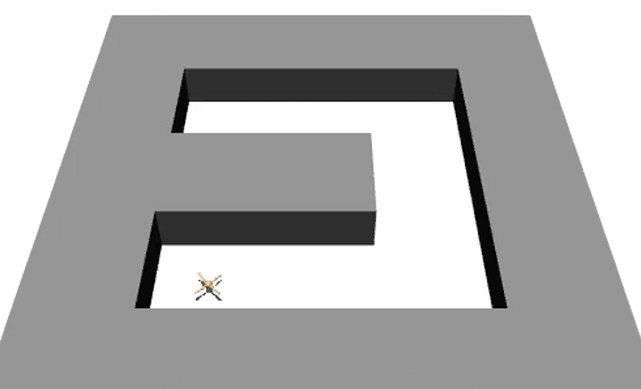}
    \subcaption{ \: Ant Maze}\label{fig:ant-b}
  \end{subfigure}
  \caption{Percentage Learning curves on Ant Fall, Ant Push, and Ant Maze comparing S3, HIRO, and HRAC. All methods are trained for 5 million environment steps. Curves report the mean performance over 5 random seeds; shaded regions denote ±1 s.e.m. Higher values indicate better task success.}
  \label{fig:ant}
\end{figure*}

A key limitation of HRL is that the high-level agent (i.e., the Manager) receives an order of magnitude fewer observations as compared to the low-level agent (i.e., the Worker). While fewer observations are what make the HRL approach tractable, the simultaneous coupling with sparse and delayed rewards creates the need for more meaningful exploration for the Manager. While recent HRL algorithms have focused on designing novel subgoal exploration methods \cite{nachum2018data, zhang2020generating, wang2025hierarchical}, we explore how incorporating dense dynamics-aware intrinsic rewards for the high-level agent can assist in the learning process.

 More specifically, the dynamics-aware intrinsic reward encourages the Manager to issue subgoals that the current Worker can execute consistently and reliably, based on its current capability -- in effect, modeling the coarse transition dynamics of the environment.
As shown in Figure \ref{fig:s3}(a), the dense intrinsic reward is provided to the manager at every temporally abstracted step (figure \ref{fig:s3-a}. Moreover, the estimation of the intrinsic reward (S3) sits atop whichever HRL algorithm is being used for the learning process. The exact process of estimating the intrinsic reward is discussed in Section \ref{Sec:MDN_Intrinsic_Reward}.

\subsection{Potential-based Reward Shaping}
To provide denser feedback to the high-level agent, we need an intrinsic reward that is meaningful and guides towards the optimal policy. 
Furthermore, for the intrinsic reward to be policy invariant, we propose shaping a potential-based intrinsic reward \cite{ng1999policy, wiewiora2003principled}. 
Following the potential-based shaping framework introduced by \citet{ng1999policy} and later generalized for state-action pairs by \citet{wiewiora2003principled}, we define a potential-based intrinsic reward over state–action pairs rather than states alone. A reward function is said to be potential-based advice if it satisfies the following equation:
\begin{equation}
\label{Eqn:PBRS reward condition - original}
    F(s,a,s',a') = \gamma \Phi_{t+c}(s', a') - \Phi_{t}(s, a)
\end{equation}
where $\gamma$ denotes the environment discount factor, and $\Phi(s,a)$ is a time-dependent bounded potential function that assigns scalar values to state-action pairs \cite{devlin2012dynamic}.
This extension allows shaping to provide more precise guidance, biasing not only which states are desirable but also which actions are promising within those states, while still preserving the set of optimal policies.
A potential-based reward shaping (PBRS) function ensures that the total reward at the end of an episode is equal to the environmental reward $R^{\text{ext}}$. 
This property arises because the shaping term, defined as the discounted difference in potential between consecutive transitions, telescopes over time. 
Specifically, the intermediate shaping rewards cancel out when summed across the entire episode, leaving only the boundary terms at the start and end of the trajectory. As a result, the addition of potential-based shaping does not alter the overall return associated with any policy and therefore does not change the optimal policy of the original MDP. Instead, it redistributes the sparse environmental reward into denser intermediate feedback that accelerates learning while preserving policy invariance.
In our HRL work, we introduce a Potential based intrinsic reward for the high-level agent that takes a similar form as Equation \ref{Eqn:PBRS reward condition - original}, but over a coarser $c$-step time horizon, i.e.,
the intrinsic reward $r_{\text{hi}}^{\text{int}}$ should satisfy the following condition:
\begin{equation}
    r_{\text{hi}}^{\text{int}} = F = \gamma\Phi_{t+c}(s_{t+c}, a_{t+c}) - \Phi_{t}(s_t, a_t)
\end{equation}

Our next step is to devise a functional form for this potential based function $\Phi(s, a)$ to evaluate the intrinsic reward for the high-level agent. This step should be performed in a manner that (a) incorporates the notion of subgoal reliability, i.e., how consistently the Worker can realize the Manager's intended subgoals given its current capability and the environment's coarse transition dynamics and (b) ensures that the resulting shaped reward preserves the optimal high-level policy while providing denser, more informative feedback that improves sample efficiency. In this view, the potential function acts as a measure of the stability or predictability of the low-level outcomes conditioned on a high-level action, guiding the Manager toward subgoals that are achievable and repeatable rather than excessive or random exploration.

In the next subsection, we propose a methodology to design this intrinsic reward potential $\Phi(s, a)$ in a manner that \emph{penalizes} the high-level agent for selecting subgoals that increase uncertainty of the coarse dynamics model of the system.

\subsection{Coarse Dynamics Dispersion Penalty as a High-level Intrinsic Reward}
The inspiration for using a coarse dynamic model to inform the learning process of the Manager-Worker team was drawn from prior organizational behavior research, which emphasizes that effective managers are both task-oriented and people-oriented, balancing the assignment of clear, achievable tasks with attention to worker capabilities \cite{yukl1992theory, fiedler1967theory}.
From a human manager-worker perspective, a key challenge for the manager is to assign subgoals that can be reliably and repeatedly achieved by a worker, to improve the team's overall performance. In our work, we draw inspiration from this approach to devise an intrinsic motivation for the high-level agent that attributes success (or failure) to the quality of subgoal assigned to the worker.
 The current state-of-the-art HRL methods \cite{dayan1992feudal, schmidhuber1992planning, kulkarni2016hierarchical, levy2017learning, li2021learning, li2021active, vezhnevets2017feudal}, generate an intrinsic reward for training the low-level agent's policy $\pi_{\text{lo}}$ based on how \emph{close} the low-level agent gets to achieving the assigned subgoal $g_t$, i.e., how accurately the low-level agent progresses towards the subgoal.
To complement the sparse environmental reward obtained by the high-level agent, we also provide intrinsic motivation that rewards the assignment of subgoals that help the low-level agent perform more reliably, i.e., increase the \emph{precision} of achieving the subgoal. It implies reduced uncertainty of the coarse dynamics, which further leads to reducing the uncertainty (and perceived nonstationarity) of the policy of the low-level agent.

The objective of this approach is to steer subgoal selection towards regions where the low-level controller can operate reliably within the partially known and stochastic environmental dynamics. This approach is particularly beneficial in environments where the task involves high-risk transitions, multi-stage dependencies (e.g., approach the block, push it aside, navigate to final goal) or structural bottlenecks, i.e., the states through which most of the successful trajectories should pass \cite{mcgovern2001automatic, menache2002q}. In such settings, reliable subgoal selection helps the high-level agent to focus exploration around strategically important regions, rather than wasting samples on subgoals that may have value, but which the current worker cannot achieve consistently. Similarly, in tasks where navigation or manipulation is organized around landmarks or waypoints \cite{kulkarni2016hierarchical, kim2021landmark}, S3 helps stabilize hierarchical coordination by subgoal selection with the low-level agent's predictable and controllable outcomes.
Conversely, we expect the advantage and need of this approach to diminish in environments where bottleneck states are not critical to success or do not exist, and where the high-level agent cannot infer meaningful structural cues about progress from observable transitions.

At a temporal scale at which manager operates, the environment's transitions can be viewed through its coarse dynamics. Coarse dynamics are induced by the low-level controller when it executes a subgoal for a fixed duration. In order to measure the reliability of the low-level agent and penalize the manager for providing highly diffcult to achieve subgoals, we introduce the notion of the coarse-grained $c$-step predictive terminal-state distribution (TSD), expressed as follows:
\begin{equation}
\label{Eqn:TSD}
    p(s_{t+c}|s_t, g_t) = \sum_{\substack{a_{t:t+c-1}\\ s_{t+1:t+c-1}}}\prod_{k=0}^{c-1}[\pi_{\text{lo}}(a_{{t+k}}|s_{t+k}, g_t)p(s_{t+k+1}|s_{t+k}, a_{t+k})]
\end{equation}
where $p(s_{t+c} | s_t, g_t)$ represents the $c$-step predictive distribution over states, i.e., for each possible state $s_{t}$, it captures the probability that the system will be in $s_{t+c}$, after $c$ time steps, starting from $s_t$ and executing the current low-level policy conditioned on subgoal $g_t$ while the environment follows its dynamics. Recent HRL research targets increasing uncertainty or novelty at the manager level, such as adding exploration bonuses \cite{li2021active, mcclinton2021hac, kim2021landmark} or prioritizing novel subgoals via learned subgoal spaces. Our work, on the other hand, provides a high-level intrinsic reward that penalizes the dispersion (variance) of the c-step predictive terminal-state distribution, thereby steering subgoal selection toward reliable, repeatable outcomes using the following potential-based reward shaping function:

\begin{equation}
\label{Eqn: Dispersion function}
    \Phi_{t+c}(s_{t+c}) = - \beta\,\Psi(\cdot|s_t, g_t)
     % \Phi_{\Psi}(s) = - \textcolor{red}{\beta\:}\mathbb{E}_{g \sim \pi_{hi}(s_{t+c}}[\Psi(\cdot|s, g)]
\end{equation}
where $\Psi$ represents a dispersion metric function defined over the $c$-step Terminal State Distribution (TSD). Several different dispersion metrics may serve this purpose, e.g., total variance (trace of covariance), generalized variance (determinant or log-determinant), largest eigenvalue (spectral radius), mean squared miss-distance to the subgoal, or Conditional Value at Risk (CVaR) of miss-distance, to name a few.
Additionally, $\beta$ represents a nonnegative scaling coefficient (intrinsic weight) for the shaping term. It ensures the intrinsic signal is strong enough to guide learning yet not so large that it overrides the environment reward. In this work, we use the following dispersion metric where
$\mathrm{tr}(\cdot)$ denotes the trace of the matrix, and $\mathrm{Cov}(\cdot)$ denotes the covariance matrix of the
$c$-step terminal state distribution::
\begin{equation*}
    \Psi(\cdot|s_t, g_t) = tr(Cov(s_{t+c}|s_t, g_t))
\end{equation*}
and the high-level PBRS intrinsic reward function can then be expressed as:
\begin{equation}
\label{eq:intrinsic-reward-final}
    \tilde{r}^{\text{hi}}_t = r^{\text{hi}}_t + \gamma \Phi_{t+c}(s_{t+c}, g_{t+c}) - \Phi_t(s_t, g_t)
\end{equation}
Since we treat the environment model as unknown, the exact c-step predictive TSD cannot be determined exactly. However, for practical purposes, it can be estimated based on $c$-sampled transitions, and subsequently utilized to evaluate the high-level reward $r_{\text{hi}}^{\text{int}}$, as discussed in the next section. At each primitive step, we relabel the subgoal using the goal-transition function $h$ as first introduced in HIRO\cite{nachum2018data} to keep the manager’s intended absolute target fixed:
\begin{equation}
    h(s_{t+1}, g_{t}, s_{t}) := s_{t} + g_{t} - s_{t+1}
\end{equation}
Algorithm \ref{alg:s3} includes the steps associated with HRL process for S3.

\begin{algorithm}
\caption{S3: Stable Subgoal Selection}
\label{alg:s3}
\begin{algorithmic}[1]
\State \textbf{Inputs:} env, period $c$, num\_episodes, ep\_length, schedule $\beta_t$
\State \textbf{Policies:} high $\pi_\text{hi}(g\mid s; \theta_{\text{hi}})$, low $\pi_{\text{lo}}(a\mid s,g; \theta_{\text{lo}})$
\State \textbf{MDN:} $\hat{p}_{\theta_{\text{MDN}}}(s_{t+c}\mid s_t,g_t)$ with mixture of $K$ Gaussians means $\mu_{\text{mix}}$, covariances $\Sigma_{\text{mix}}$
\State \textbf{Replay buffers:} $\mathcal{B}_\text{hi} \gets \emptyset$, $\mathcal{B}_\text{lo} \gets \emptyset$, $\mathcal{B}_{\text{MDN}} \gets \emptyset$

\For{$\text{ep}=1$ \textbf{to} num\_episodes}
  \State\textbf{Initialize:} $s_0 \gets \mathrm{env.reset}()$,  $g_0 \gets \pi_\text{hi}(s_0)$
  \State $\Phi_0 \gets 0$, $a_0 \gets \pi_\text{lo}(s_0, g_0)$

  \For{$t=1$ \textbf{to} ep\_length}
    \State $s_t,\, r_t^{\text{env}},\, \text{done} \gets \mathrm{env.step}(a_{t-1})$
    \State $r_t^{\text{lo}} \gets -\left\lVert (s_{t-1}+g_{t-1})-s_t \right\rVert_2$
    \State push $(s_{t-1}, g_{t-1}, a_{t-1}, r_t^{\text{lo}}, s_t, \text{done})$ into $\mathcal{B}_\text{lo}$

    \If{$\text{done}$ \textbf{or} $(t \bmod c = 0)$}
      \State $g_t \gets \pi_\text{hi}(s_t)$
      \State $(\mu_\text{mix}, \Sigma_\text{mix}) \gets \hat{p}_{\theta_\text{MDN}}(\cdot \mid s_{t-c}, g_{t-c})$
      \State $\mathrm{var}_t \gets \mathrm{tr}\!\left(\Sigma_\text{mix}\right)$
      \State $\Phi_t \gets -\beta_t \cdot \mathrm{var}_t$
	\State $\tilde r^{\text{hi}} \gets r^{\text{env}}_{t-c+1:t} + \Phi_t - \Phi_{t-c}$
      \State push $(s_{t-c}, g_{t-c}, s_t)$ into $\mathcal{B}_\text{MDN}$
      \State push $(s_{t-c}, g_{t-c}, \tilde r^{\text{hi}}, s_t, \text{done})$ into $\mathcal{B}_\text{hi}$
      \State \textbf{UpdateMDN}$(\mathcal{B}_\text{MDN}; \theta_\text{MDN})$
      \State \textbf{UpdateHigh}$(\mathcal{B}_\text{hi}; \theta_\text{hi})$
    \Else
     \State $g_t \gets h(s_t,\, g_{t-1},\, s_{t-1})$
    \EndIf

    \State \textbf{UpdateLow}$(\mathcal{B}_\text{lo}; \theta_\text{lo})$
    \State $a_t \gets \pi_\text{lo}(s_t, g_t)$
    \If{$\text{done}$} \textbf{break} \EndIf
  \EndFor
\EndFor
\end{algorithmic}
\end{algorithm}

%%%%%%%%%%%%%%%%%%%%%%%%%%%%%%%%%%%%%%%%%%%%%%%%%%%%%%%%%%%%%%%%%%%%%%%%
\section{Estimation of Intrinsic Reward}
\label{Sec:MDN_Intrinsic_Reward}

As discussed earlier, the high-level agent learning explicitly depends on the dense dynamics-aware intrinsic reward it receives. However, each episode only provides a single state-space trajectory based on the current policy, making it impossible to obtain the statistics of the $c$-step terminal state distribution. In order to provide the high-level agent with an uncertainty estimate pertaining to the TSD, we rely on a replay buffer $\mathcal{B}_{\text{S3}}$ that records the following quantities in each episode: the initial state $s_t$ of the low-level agent at time $t$, the goal $g_t$ assigned by the high-level agent at time $t$, and the terminal state $s_{t+c}$ of the low-level agent at time $t+c$, i.e., the end of the $c$-step time interval, after which a new subgoal is assigned by the high-level agent. As described in Figure \ref{fig:s3-b}, we sample observations $s_t$ and $s_{t+c}$ from the environment, $g_t$ from high-level policy $\pi_{\text{hi}}$ and train our coarse predictive model with these transition tuples. Thus, for any given input feature $\{s_t, g_t\}$, a user-defined supervised learning model $f(s_t, g_t)$ learns to predict the output label $\{s_{t+c}\}$. At the time of training the high-level agent, we query the trained model $f^*(s_t, g_t)$ to obtain the statistics of $s_{t+c}$ via a dispersion metric, which is subsequently used to provide the dense uncertainty-based high-level intrinsic reward (Equation \ref{eq:intrinsic-reward-final}). 
 
The machine learning model $f(s_t, g_t)$ we use in our work is based on a Mixture Density Network (MDN) \cite{bishop1994mixture}, and is used to model the (potentially) multi-modal dynamics $P(s_{t+c}|s_t, g_t)$ as a mixture of multiple Gaussians. Instead of blurring divergent outcomes into one mean, MDN represents distinct terminal modes with its each mean and covariance (discussed below):
\begin{equation}
    \hat{p}_{\theta_{\text{MDN}}}(s_{t+c}|s_t, g_t) = \sum_{k=1}^K \alpha_k(s_t, g_t) \mathcal{N}(s_{t+c};\mu_k(s_t, g_t), \Sigma_k(s_t, g_t)) 
\end{equation}
% \ssuggestion{where $\alpha_k$ represents..., $\mu_k$ etc. [we need this for each term that has not been introduced before...]} 
In an MDN, the coarse dynamics predictive distribution $p(s_{t+c}|s_t, g_t)$ $\approx \hat{p}_{\theta_{\text{MDN}}}(s_{t+c}|s_t,g_t)$ is modeled as a $K$-component Gaussian mixture with weights $\alpha_k(s_t,g_t)$, means $\mu_k(s_t,g_t)$ and covariances $\Sigma_k(s_t,g_t)$. Each mean $\mu_k(s_t,g_t)$ encodes a distinct terminal outcome of the macro action.
% e.g., straight push to clear aisle, yaw-left into the wall, or yaw-right drift. 
The weights $\alpha_k(s_t,g_t)$ represent the the probability of each scenario under current state $s_t$ and subgoal $g_t$. The covariance matrices $\Sigma_k$ capture the directional spread of landing locations around each $\mu_k$, quantifying the uncertainty induced by low-level controller stochasticity and unmodeled dynamics.

Jointly optimizing the manager and the worker policies is intrinsically non-stationary, and layering an MDN on top can in principle amplify that instability. However, the evolving policy furnishes a useful curriculum: early iterations feed the MDN simple, high-variance trajectories, while later policies contribute sharper, more targeted behaviors, allowing the mixture to expand its support progressively rather than overfitting a snapshot of the given policies. 

The predictive uncertainty, i.e., covariance matrices conditioned on $s_t$ and $g_t$, is split into $K$ Gaussian distributions. The total predictive covariance decomposes into within-mode and between-mode terms.
\begin{equation}
    \Sigma_{\text{mix}}
    \;=\; \underbrace{\sum_{k=1}^{K} \omega_k\, \Sigma_k}_{\text{within-mode}}
    \;+\;
    \underbrace{\sum_{k=1}^{K} \omega_k\, (\mu_k - \mu)(\mu_k - \mu)^\top}_{\text{between-mode}}
\end{equation}

Within-mode quantifies variability inside each scenario and between-mode measures ambiguity over which outcome will occur. 
Together, a smaller value of $\Sigma_{\text{mix}}$ signifies a more predictable outcome from the low-level agent for the current state and subgoal. It further suggests that each macro-action would terminate near its mode mean. This gives the Manager the ability to plan with predictable behavior of the Worker, hence, improving subgoal feasibility and sample efficiency.

To preserve the optimality of the underlying solution, we apply potential-based reward shaping \cite{wiewiora2003principled}. The shaping potential is defined over high-level state pairs and penalizes dispersion in the MDN’s coarse dynamics distribution:
\begin{equation}
    \Phi_t(s_t, g_t) \;=\; -\beta\,\Psi\!\big(\Sigma_{\text{mix}}(s_{t+c}\mid s_t, g_t)\big),
\end{equation}
We tune the intrinsic-reward coefficient $\beta$ to inject a sufficiently large shaping signal to drive joint estimation of the coarse dynamics and value model, while constraining $\beta$ so that the intrinsic term remains a lower-order perturbation relative to the extrinsic reward.

%%%%%%%%%%%%%%%%%%%%%%%%%%%%%%%%%%%%%%%%%%%%%%%%%%%%%%%%%%%%%%%%%%%%%%%%
\section{Empirical Study}
\label{Sec:Empirical Study}

\begin{table}[t]
	\caption{We compare S3 against HIRO and HRAC. Each method is trained for 5M environment steps under five random seeds. We report mean success rate (percentage of episodes that reach the final goal) $\pm$ standard error of the mean (SEM) across seeds.}
	\label{tab:all-env-results}
	\begin{tabular}{r|ccc}\toprule
		\textit{Environment} & \textit{S3} & \textit{HIRO} & \textit{HRAC} \\ \midrule
		Ant Fall & \textbf{0.374$\pm$0.054} & 0.059$\pm$0.059 & 0.000$\pm$0.000\\
		Ant Push & \textbf{0.186$\pm$0.107} & 0.134$\pm$0.101 & 0.180$\pm$0.180 \\
		Ant Maze & 0.827$\pm$0.024 & 0.790$\pm$0.067 & \textbf{0.832$\pm$0.039}\\ \bottomrule
	\end{tabular}
\end{table}
\begin{figure}[t]
    \centering
    \includegraphics[width=\linewidth]{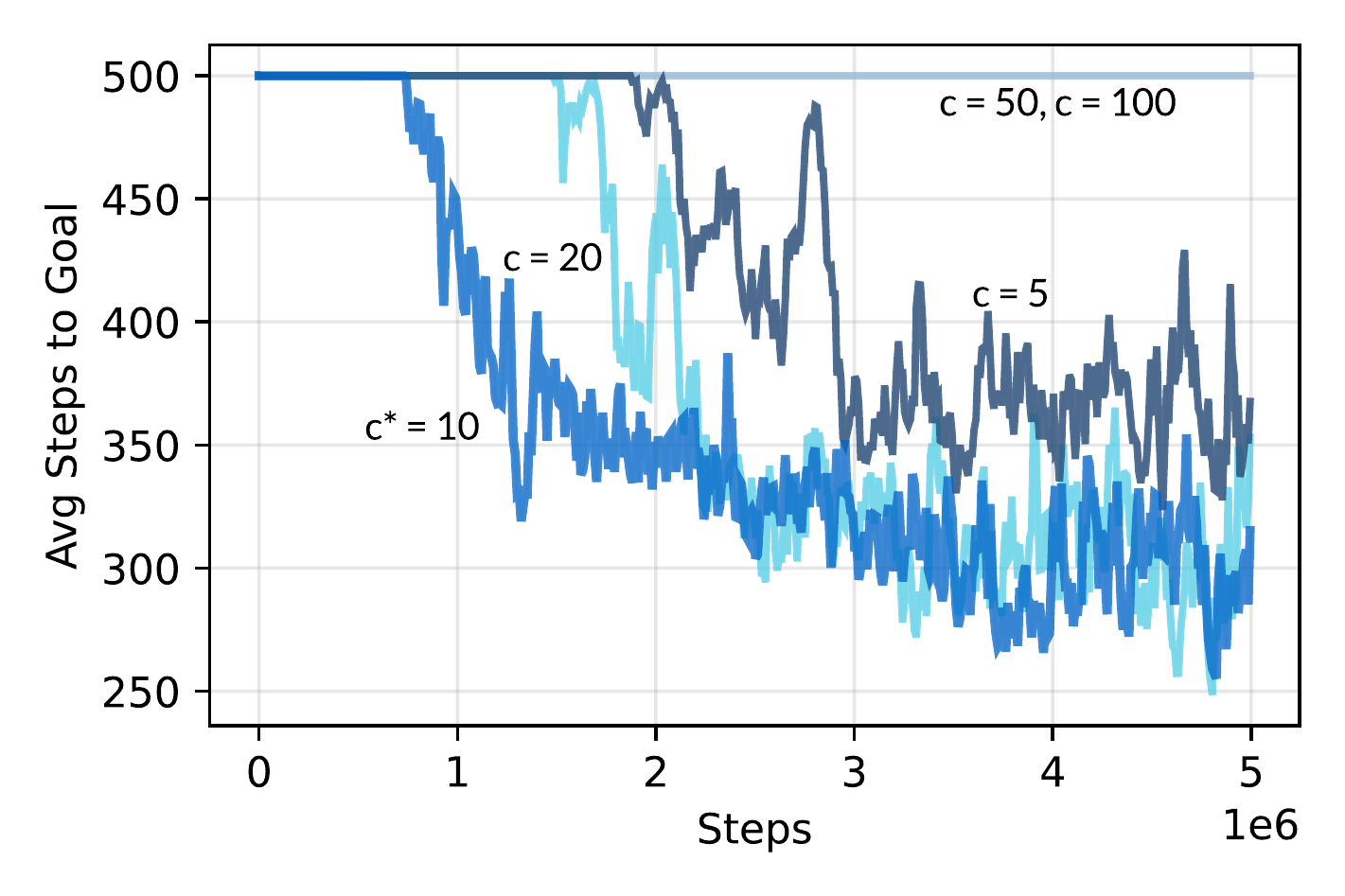}
    \caption{S3 learning curves under different Manager horizons. X-axis shows environment steps and Y-axis represents the average steps to finish (lower is better). Each curve shows periodic evaluation of a fixed $c$ (Manager horizon) compared to our choice of Manager horizon interval $c^*=10$ while training. \vspace{-1em}}
    \label{fig:horizon-avg-steps}
\end{figure}
\begin{figure*}[!t]
    \centering
    \includegraphics[width=0.3\linewidth]{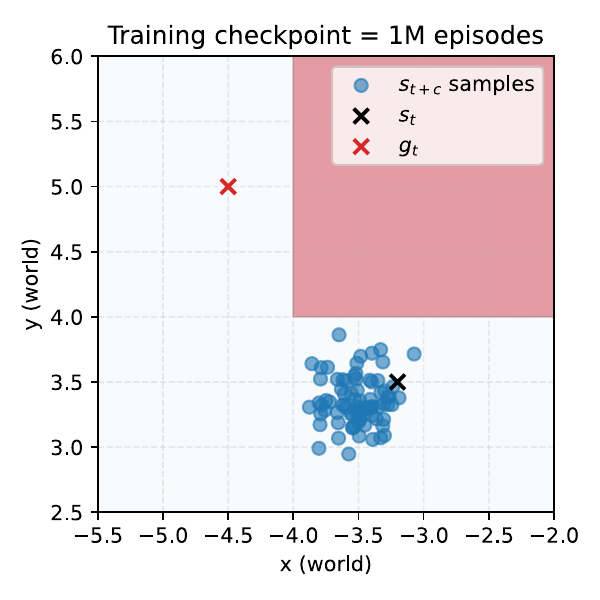}
    \includegraphics[width=0.3\linewidth]{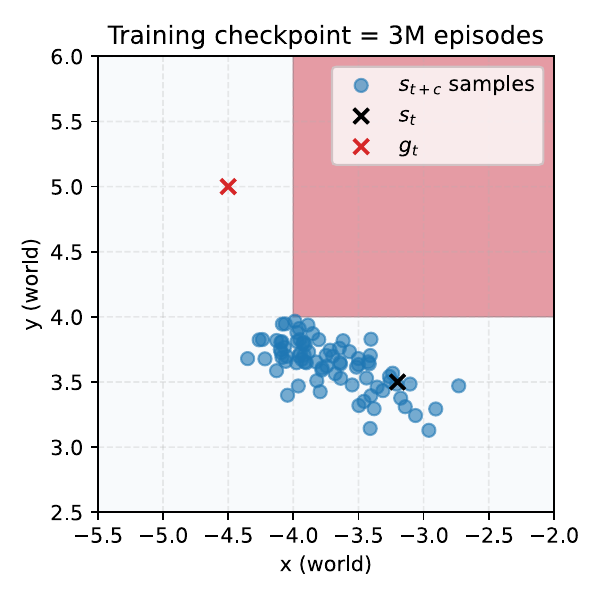}
    \includegraphics[width=0.3\linewidth]{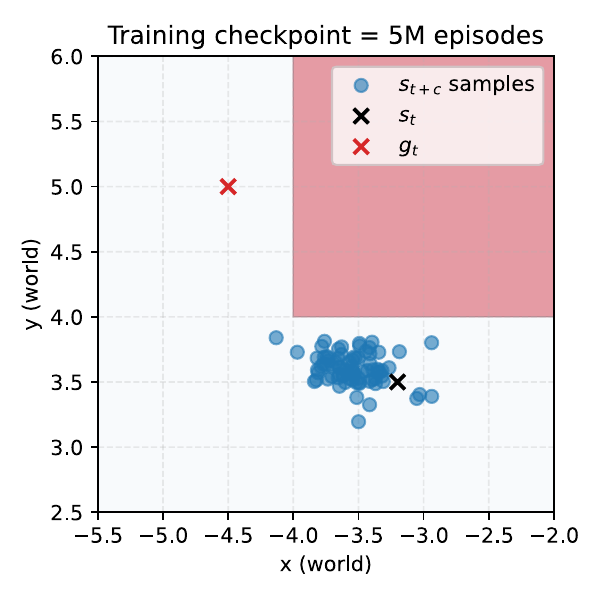}
    \caption{Scatter plot designed to show the spread of worker capability at different points in the training. Blue dots are terminal states $s_{t+c}$, black X is current state $s_t$, red X is assigned subgoal $g_t$, and the red-shaded quadrant is the hazardous zone where a push from the south side would irreversibly block the goal. The above checkpoints are showcased for Ant Push environment}
    \label{fig:worker-scatter}
\end{figure*}

We evaluate S3 on three standard MuJoCo continuous-control benchmarks (Ant Maze, Ant Push, and Ant Fall)\cite{todorov2012physics} using the quadruped Ant agent. We choose these tasks for three reasons: (i) they support temporal abstraction, allowing a high-level policy to issue goal-like subgoals while a low-level controller handles locomotion; (ii) success requires multi-stage behavior, e.g., navigating to intermediate way-points and interacting with objects before reaching the final goal; and (iii) these environments are widely adopted HRL benchmarks, hence, making standardized evaluation and fair comparison with prior work simpler.
\subsection{Environment Setup}
Ant is a quadruped agent with two actuated joints per leg (8 actuators in total). The action space consists of continuous joint torque commands $a_t \in [-1, 1]^8$ clipped by simulator limits. The observations includes joint positions/velocities, base pose and velocities, and task variables (e.g., torso $(x,y)$, goal and object poses).
Figure \ref{fig:ant} shows the three different tasks discussed below:

\noindent\textbf{Ant Maze.} Navigation through narrow U-shaped/maze-like corridors to a distant goal specified in workspace coordinates (goal-conditioned RL). Rewards are typically sparse (success on entering the goal region). It requires long-horizon planning and effective subgoal setting.

\noindent\textbf{Ant Push.} A movable block obstructs the path between start and goal. The agent must push the block in an appropriate direction to clear a passage before navigating to the goal. Greedy go-to-goal behavior will fail, leading to incorrect pushes further blocking access.

\noindent\textbf{Ant Fall.} There are two platforms separated by a gap. The agent must push a nearby block to form a bridge, then traverse it to reach the goal. Failed manipulation or misplacement of the block leads to irreversible failure (e.g., block falls into the gap).

\subsection{Comparative Analysis}

In Ant Push, the agent must first align and push a movable block to clear a passage before navigation can proceed. In Ant Fall, the agent must set up a precise approach and drop the block in the gap before continuing, where small approach differences lead to sharply different post-contact outcomes. These tasks induce heteroscedastic, multi-modal coarse dynamics, where distinct contact outcomes lead to sharply different futures. 

As observed in Figure \ref{fig:ant}, S3 achieves significantly higher success rates on Ant Fall than the baselines.
It also outperforms baselines on Ant Push, albeit with a smaller margin. 
This gain stems from Ant Fall’s stronger contact-induced multi-modality (e.g., precise block placement vs. jam), where other baseline models blur subgoal outcomes, while the MDN preserves distinct modes with calibrated probabilities. 
Both AntPush and AntFall impose long-horizon preconditions (clear the block; place the block), where compounding (or not accounting for) model error typically hurts baseline approaches; S3’s coarse dynamics are concentrated within each scenario, stabilizing subgoal selection and execution over extended horizons. 
S3 attains comparable performance on AntMaze to the baselines. 
Here, accurately modeling coarse dynamics offers limited marginal benefit: navigation is dominated by predictable locomotion and geometric path-finding, so planning quality is determined more by subgoal decomposition and low-level exploration than by coarse dynamics dispersion. 
Hence, we can safely conclude that S3 is useful in tasks involving substantial physical interaction and long-horizon preconditions, where outcome variability is strongly state-dependent and a coarse dynamics model based intrinsic reward materially improves subgoal selection and execution in the manager.

\subsection{Qualitative Analysis}
\label{sec:empirics-ablation}
\begin{figure*}[!t]
    \centering
    \begin{subfigure}[t]{0.33\textwidth}    
            \centering
            \includegraphics[width=\linewidth]{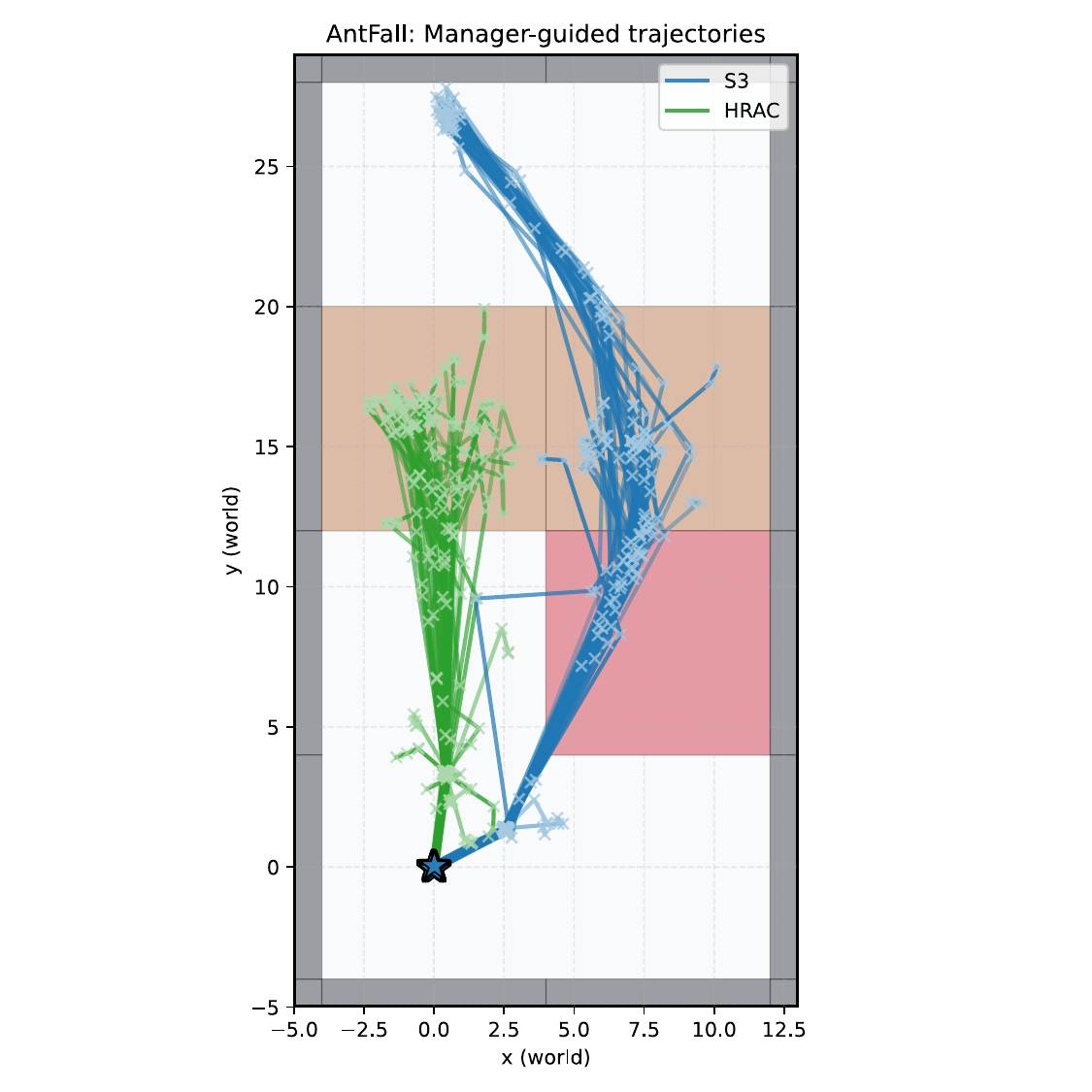}
            \subcaption{\: Ant Fall}
            \label{fig:traj-manager-fall}
    \end{subfigure}
    \begin{subfigure}[t]{0.33\textwidth}
        \centering
            \includegraphics[width=\linewidth]{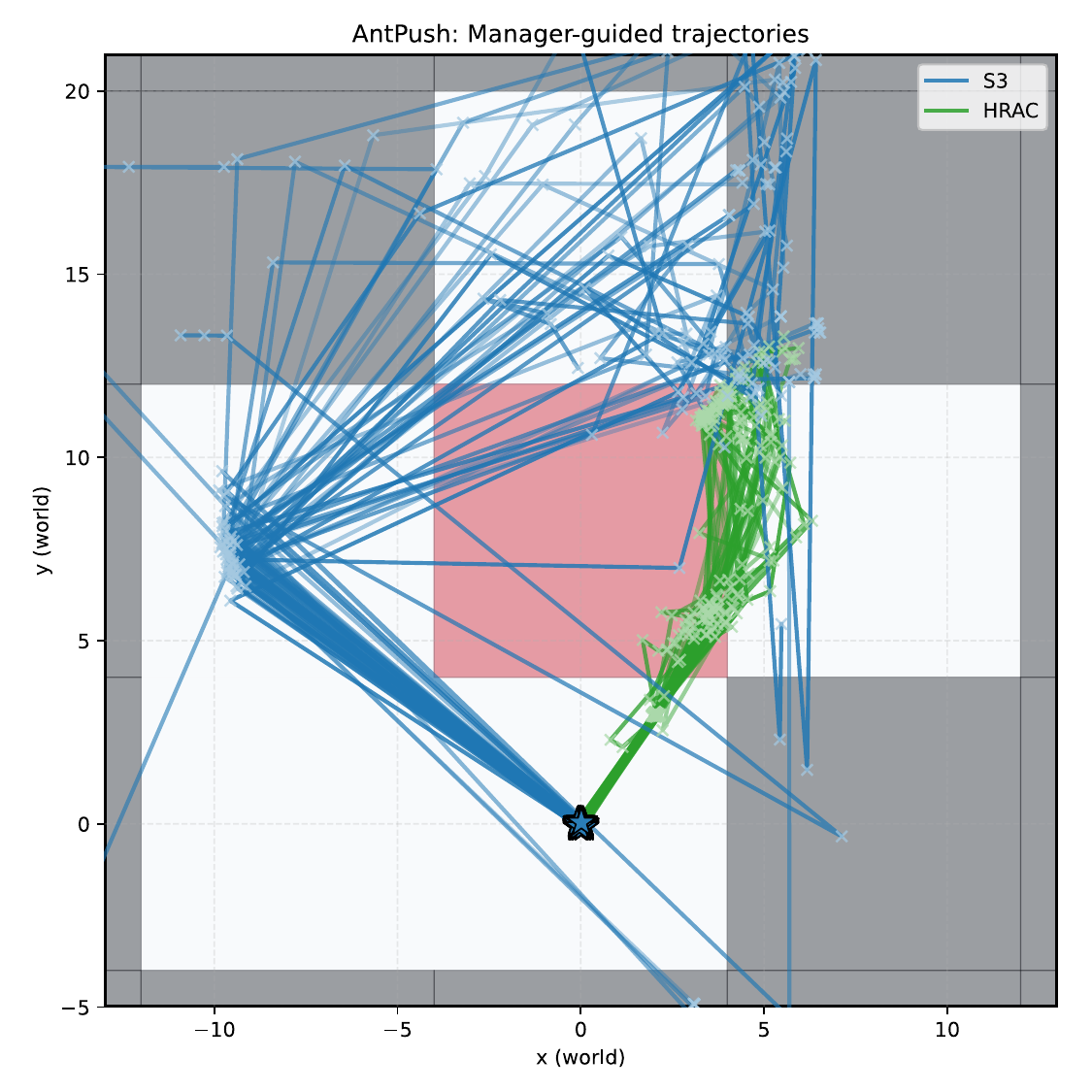}
        \subcaption{\: Ant Push}
        \label{fig:traj-manager-push}
    \end{subfigure}
    \begin{subfigure}[t]{0.33\textwidth}
        \centering
            \includegraphics[width=\linewidth]{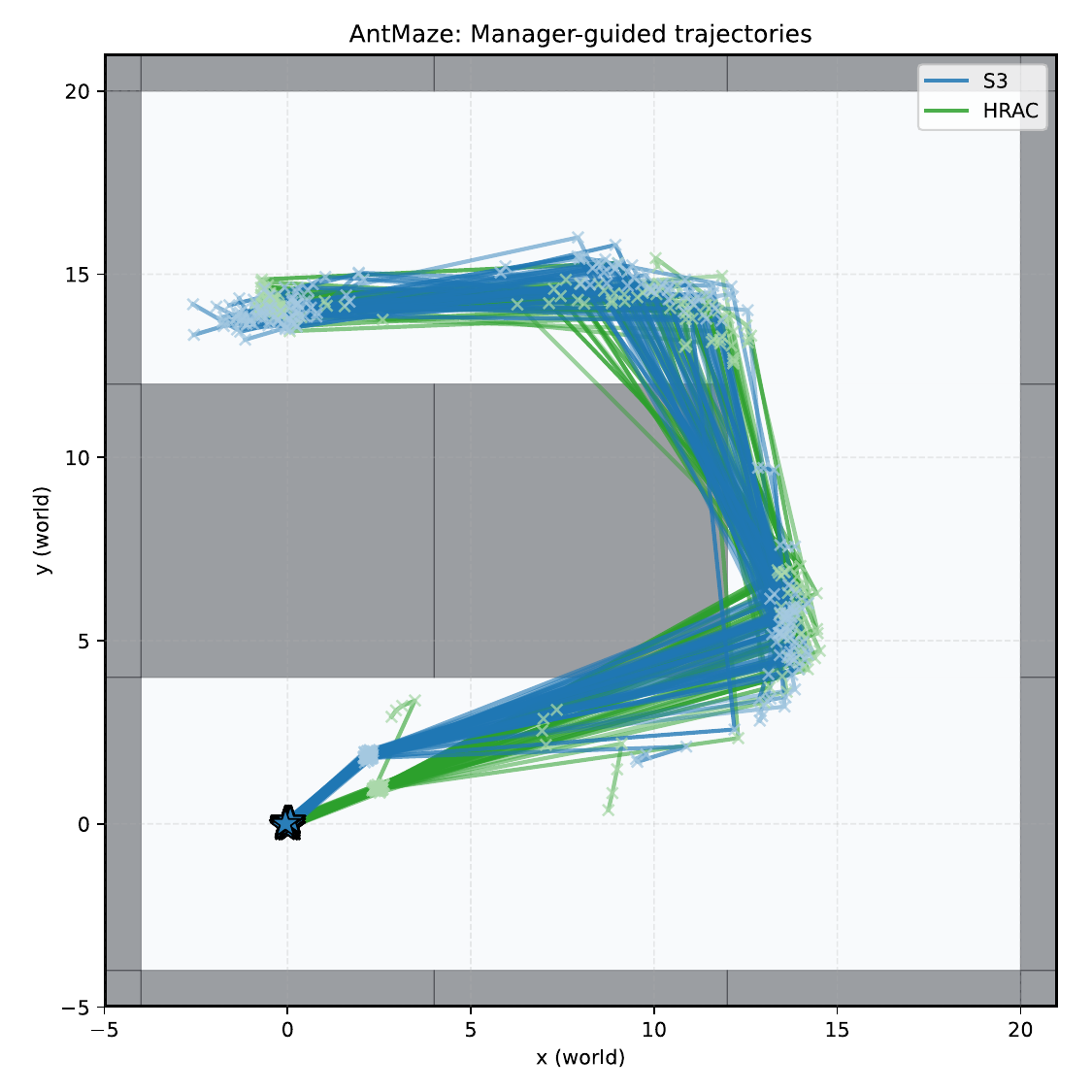}
        \subcaption{\: Ant Maze}
        \label{fig:traj-manager-maze}
    \end{subfigure}
    \caption{Manager-guided trajectories for S3 (blue) and HRAC (green) over 50 evaluation runs in (a) Ant Fall, (b) Ant Push, and (c) Ant Maze. Each trajectory corresponds to a single episode, with 'x' markers plotted at every 10th high-level subgoal to reduce visual clutter and highlight the dominant route chosen by the manager over the course of the run.}
    \label{fig:traj-manager}
\end{figure*}

We hypothesize that more frequent subgoal selection tighten landing prediction by limiting rollout drift, while less frequent subgoals trade that predictability for stronger temporal abstraction and longer-horizon commitment. To evaluate S3 for sensitivity towards varying temporal horizons, we sweep across temporal abstraction horizon $c\in\{5, 10, 20, 50, 100\}$ with identical configurations on Ant Maze task. The curve for S3, which is trained with manager horizon $c=10$ attains the lowest average steps to complete the task (Figure \ref{fig:horizon-avg-steps}).
With identical training hyperparameters, S3, which has manager horizon c=10 minimizes average steps, marking a “stability region” where the high-level proposals are (i) ambitious enough to generate measurable progress per subgoal and (ii) still within the worker’s controllable horizon so that execution errors do not accumulate unchecked.
While $c=5$ under-utilizes the low-level agent's capability to generate complex trajectory for a given subgoal. For $c= 20$, despite the increasing multi-modality, longer horizon shows comparable learning. For $c\geq50$, the degrading calibration between high-level policy, low-level policy and intrinsic reward estimator becomes unstable for learning.  

In Figure \ref{fig:worker-scatter}, we illustrate learning coarse dynamics in Ant Push environment. The red quadrant marks a hazardous zone: if the agent pushes the movable block from the south, the block enters an irreversible configuration and the final task goal becomes unreachable. Each panel plots the manager's current state $s_t$ in a black cross, the assigned subgoal $g_t$ in red cross, and a cloud of sampled terminal states  $s_{t+c}$ generated from the worker's policy.

Over training, by 1M episodes samples remain fairly dispersed and frequently overlap the red zone, indicating that the model and policy have not yet internalized the risk. Around 3M episodes, the terminal-state cloud shifts toward the subgoal and becomes sharper. Also, the number of samples entering the danger zone becomes significantly less. This may indicate that the coarse dynamics model now captures the causal consequence of pushing near the block. By 5M episodes, the predictions form a concentrated cluster that stays clear of the red region while trying to achieve $g_t$.

Consequently, the sampled terminal states $s_{t+c}$ increasingly concentrate toward the commanded subgoal $g_t$ over training, indicating that the worker continues to execute meaningful progress toward manager proposals rather than collapsing to negligible motion around $s_t$. Moreover, the reduction of mass entering the hazardous quadrant suggests that shaping the manager toward lower-dispersion subgoals selectively avoids risky macro-transitions while preserving low-level learning.
Figure \ref{fig:traj-manager} visualizes manager-only trajectories formed by consecutive subgoals $g_t$.
In Ant Fall (Figure \ref{fig:traj-manager-fall}), HRAC’s manager subgoals are loosely goal-directed but scattered, indicating no consistent strategy for using the block to traverse the gap. S3 concentrates subgoals around the bridge-crossing region, reflecting an explicit focus on reliable crossing where stochasticity can cause the agent to fall off the bridge. 
In Ant Push (Figure \ref{fig:traj-manager-push}), S3’s manager subgoals look scattered because it repeatedly re-plans around the movable block. When the coarse dynamics model predicts high uncertainty, where a small push can irreversibly block the goal, the shaping term penalizes those subgoals and the manager switches to safer targets, producing criss-crossing waypoints across runs. Despite this, the realized trajectories are tightly concentrated along a smooth left-hand arc that avoids the block and reaches the goal.
S3 produces a narrow, overlapping pathway that are tightly aligned with the task direction, whereas HRAC exhibits noticeably thicker bundles of trajectories in \ref{fig:traj-manager-maze}. 
%%%%%%%%%%%%%%%%%%%%%%%%%%%%%%%%%%%%%%%%%%%%%%%%%%%%%%%%%%%%%%%%%%%%%%%%
\section{Conclusion}
We address the challenges in learning hierarchical policies like high-level exploration and credit assignment, inherent to tasks with long-horizon dependencies. Our presented approach leverages the ideas and bridges the gap between model-based HRL, potential-based shaping, and information–theoretic intrinsic motivation, to create a method to enable the high-level agent to build a coarse model of environment dynamics. By leveraging coarse dynamics of subgoal outcomes, we introduce an intrinsic reward that quantifies subgoal reliability while preserving the optimality of the high-level policy. This combination provides dense feedback to the Manager, mitigates the non-stationarity arising from the evolving Worker, and yields stable hierarchical learning in sparse-reward, long-horizon environments.
Empirically, we observe superior performance compared to HRL baselines in long-horizon dependency and continuous control environments with bottleneck states. S3 is most beneficial near bottlenecks and irreversible transitions, where small execution variance can push the system into unrecoverable regions. Through further investigation, we observe that S3 supports and learns from long manager interval. 
As the primary contribution of our work is the design of the dynamics-aware high-level intrinsic reward, our approach is generalizable to other HRL algorithms.
%%%%%%%%%%%%%%%%%%%%%%%%%%%%%%%%%%%%%%%%%%%%%%%%%%%%%%%%%%%%%%%%%%%%%%%%
%%% The acknowledgments section is defined using the "acks" environment
%%% (rather than an unnumbered section). The use of this environment 
%%% ensures the proper identification of the section in the article 
%%% metadata as well as the consistent spelling of the heading.
\begin{acks}
This research has been funded by the Office of Naval Research (N00014-23-1-2744 and N00014-24-1-2634)
\end{acks}

%%%%%%%%%%%%%%%%%%%%%%%%%%%%%%%%%%%%%%%%%%%%%%%%%%%%%%%%%%%%%%%%%%%%%%%%

%%% The next two lines define, first, the bibliography style to be 
%%% applied, and, second, the bibliography file to be used.

\bibliographystyle{ACM-Reference-Format} 
\bibliography{references}

@inproceedings{ng1999policy,
  title={Policy invariance under reward transformations: Theory and application to reward shaping},
  author={Ng, Andrew Y and Harada, Daishi and Russell, Stuart},
  booktitle={Icml},
  volume={99},
  pages={278--287},
  year={1999},
  organization={Citeseer}
}

@article{haeri2022reward,
  title={Reward-sharing relational networks in multi-agent reinforcement learning as a framework for emergent behavior},
  author={Haeri, Hossein and Ahmadzadeh, Reza and Jerath, Kshitij},
  journal={arXiv preprint arXiv:2207.05886},
  year={2022}
}

@article{barto2003recent,
  title={Recent advances in hierarchical reinforcement learning},
  author={Barto, Andrew G and Mahadevan, Sridhar},
  journal={Discrete event dynamic systems},
  volume={13},
  number={4},
  pages={341--379},
  year={2003},
  publisher={Springer}
}

@article{nachum2018data,
  title={Data-efficient hierarchical reinforcement learning},
  author={Nachum, Ofir and Gu, Shixiang Shane and Lee, Honglak and Levine, Sergey},
  journal={Advances in neural information processing systems},
  volume={31},
  year={2018}
}

@article{zhang2020generating,
  title={Generating adjacency-constrained subgoals in hierarchical reinforcement learning},
  author={Zhang, Tianren and Guo, Shangqi and Tan, Tian and Hu, Xiaolin and Chen, Feng},
  journal={Advances in neural information processing systems},
  volume={33},
  pages={21579--21590},
  year={2020}
}

@unpublished{bishop1994mixture,
       publisher = {Aston University},
         address = {Birmingham},
          author = {Christopher M. Bishop},
            year = {1994},
            type = {Technical Report},
          series = {NCRG},
           title = {Mixture density networks}
}

@inproceedings{wiewiora2003principled,
  title={Principled methods for advising reinforcement learning agents},
  author={Wiewiora, Eric and Cottrell, Garrison W and Elkan, Charles},
  booktitle={Proceedings of the 20th international conference on machine learning (ICML-03)},
  pages={792--799},
  year={2003}
}

@article{andrychowicz2017hindsight,
  title={Hindsight experience replay},
  author={Andrychowicz, Marcin and Wolski, Filip and Ray, Alex and Schneider, Jonas and Fong, Rachel and Welinder, Peter and McGrew, Bob and Tobin, Josh and Abbeel, Pieter  and Zaremba, Wojciech},
  journal={Advances in neural information processing systems},
  volume={30},
  year={2017}
}

@article{kim2021landmark,
  title={Landmark-guided subgoal generation in hierarchical reinforcement learning},
  author={Kim, Junsu and Seo, Younggyo and Shin, Jinwoo},
  journal={Advances in neural information processing systems},
  volume={34},
  pages={28336--28349},
  year={2021}
}

@article{wang2025hierarchical,
  title={Hierarchical Reinforcement Learning with Uncertainty-Guided Diffusional Subgoals},
  author={Wang, Vivienne Huiling and Wang, Tinghuai and Pajarinen, Joni},
  journal={arXiv preprint arXiv:2505.21750},
  year={2025}
}

@article{mcclinton2021hac,
  title={Hac explore: Accelerating exploration with hierarchical reinforcement learning},
  author={McClinton, Willie and Levy, Andrew and Konidaris, George},
  journal={arXiv preprint arXiv:2108.05872},
  year={2021}
}

@article{nachum2019multi,
  title={Multi-agent manipulation via locomotion using hierarchical sim2real},
  author={Nachum, Ofir and Ahn, Michael and Ponte, Hugo and Gu, Shixiang and Kumar, Vikash},
  journal={arXiv preprint arXiv:1908.05224},
  year={2019}
}

@article{gao2024hierarchical,
  title={Hierarchical reinforcement learning from demonstration via reachability-based reward shaping},
  author={Gao, Xiaozhu and Liu, Jinhui and Wan, Bo and An, Lingling},
  journal={Neural Processing Letters},
  volume={56},
  number={3},
  pages={184},
  year={2024},
  publisher={Springer}
}

@article{yukl1992theory,
  title={Theory and research on leadership in organizations.},
  author={Yukl, Gary and Van Fleet, David D},
  year={1992},
  publisher={Consulting Psychologists Press}
}

@article{fiedler1967theory,
  title={A THEORY OF LEADERSHIP EFFECTIVENESS. MCGRAW-HILL SERIES IN MANAGEMENT.},
  author={Fiedler, Fred E},
  year={1967},
  publisher={ERIC}
}

@INPROCEEDINGS{todorov2012physics,
  author={Todorov, Emanuel and Erez, Tom and Tassa, Yuval},
  booktitle={2012 IEEE/RSJ International Conference on Intelligent Robots and Systems}, 
  title={MuJoCo: A physics engine for model-based control}, 
  year={2012},
  volume={},
  number={},
  pages={5026-5033},
  doi={10.1109/IROS.2012.6386109}}

@inproceedings{dayan1992feudal,
 author = {Dayan, Peter and Hinton, Geoffrey E},
 booktitle = {Advances in Neural Information Processing Systems},
 editor = {S. Hanson and J. Cowan and C. Giles},
 pages = {},
 publisher = {Morgan-Kaufmann},
 title = {Feudal Reinforcement Learning},
 url = {https://proceedings.neurips.cc/paper_files/paper/1992/file/d14220ee66aeec73c49038385428ec4c-Paper.pdf},
 volume = {5},
 year = {1992}
}

@inproceedings{schmidhuber1992planning,
  title={Planning simple trajectories using neural subgoal generators},
  author={Schmidhuber, J{\"u}rgen and Wahnsiedler, Reiner},
  booktitle={Proceedings of the 2nd International Conference on Simulation of Adaptive Behavior},
  pages={196--202},
  year={1992}
}

@article{kulkarni2016hierarchical,
  title={Hierarchical deep reinforcement learning: Integrating temporal abstraction and intrinsic motivation},
  author={Kulkarni, Tejas D and Narasimhan, Karthik and Saeedi, Ardavan and Tenenbaum, Josh},
  journal={Advances in neural information processing systems},
  volume={29},
  year={2016}
}

@article{levy2017learning,
  title={Learning multi-level hierarchies with hindsight},
  author={Levy, Andrew and Konidaris, George and Platt, Robert and Saenko, Kate},
  journal={arXiv preprint arXiv:1712.00948},
  year={2017}
}

@inproceedings{li2021learning,
  title={Learning subgoal representations with slow dynamics},
  author={Li, Siyuan and Zheng, Lulu and Wang, Jianhao and Zhang, Chongjie},
  booktitle={International Conference on Learning Representations},
  year={2021}
}

@article{li2021active,
  title={Active hierarchical exploration with stable subgoal representation learning},
  author={Li, Siyuan and Zhang, Jin and Wang, Jianhao and Yu, Yang and Zhang, Chongjie},
  journal={arXiv preprint arXiv:2105.14750},
  year={2021}
}

@inproceedings{vezhnevets2017feudal,
  title={Feudal networks for hierarchical reinforcement learning},
  author={Vezhnevets, Alexander Sasha and Osindero, Simon and Schaul, Tom and Heess, Nicolas and Jaderberg, Max and Silver, David and Kavukcuoglu, Koray},
  booktitle={International conference on machine learning},
  pages={3540--3549},
  year={2017},
  organization={PMLR}
}

@article{eysenbach2018diayn,
  title={Diversity is all you need: Learning skills without a reward function},
  author={Eysenbach, Benjamin and Gupta, Abhishek and Ibarz, Julian and Levine, Sergey},
  journal={arXiv preprint arXiv:1802.06070},
  year={2018}
}

@inproceedings{klyubin2005empowerment,
  title={Empowerment: A universal agent-centric measure of control},
  author={Klyubin, Alexander S and Polani, Daniel and Nehaniv, Chrystopher L},
  booktitle={2005 ieee congress on evolutionary computation},
  volume={1},
  pages={128--135},
  year={2005},
  organization={IEEE}
}

@article{gregor2016variational,
  title={Variational intrinsic control},
  author={Gregor, Karol and Rezende, Danilo Jimenez and Wierstra, Daan},
  journal={arXiv preprint arXiv:1611.07507},
  year={2016}
}

@article{sutton1999between,
  title={Between MDPs and semi-MDPs: A framework for temporal abstraction in reinforcement learning},
  author={Sutton, Richard S and Precup, Doina and Singh, Satinder},
  journal={Artificial intelligence},
  volume={112},
  number={1-2},
  pages={181--211},
  year={1999},
  publisher={Elsevier}
}

@article{mcgovern2001automatic,
  title={Automatic discovery of subgoals in reinforcement learning using diverse density},
  author={McGovern, Amy and Barto, Andrew G},
  year={2001}
}

@inproceedings{menache2002q,
  title={Q-cut—dynamic discovery of sub-goals in reinforcement learning},
  author={Menache, Ishai and Mannor, Shie and Shimkin, Nahum},
  booktitle={European conference on machine learning},
  pages={295--306},
  year={2002},
  organization={Springer}
}

@inproceedings{brys2015policy,
author = {Brys, Tim and Harutyunyan, Anna and Taylor, Matthew E. and Now\'{e}, Ann},
title = {Policy Transfer using Reward Shaping},
year = {2015},
isbn = {9781450334136},
publisher = {International Foundation for Autonomous Agents and Multiagent Systems},
address = {Richland, SC},
booktitle = {Proceedings of the 2015 International Conference on Autonomous Agents and Multiagent Systems},
pages = {181–188},
numpages = {8},
location = {Istanbul, Turkey},
series = {AAMAS '15}
}

@inproceedings{devlin2014potential,
  title={Potential-based difference rewards for multiagent reinforcement learning},
  author={Devlin, Sam and Yliniemi, Logan and Kudenko, Daniel and Tumer, Kagan},
  booktitle={Proceedings of the 2014 international conference on Autonomous agents and multi-agent systems},
  pages={165--172},
  year={2014}
}

@article{watanabe2022shiro,
  title={{SHIRO}: Soft hierarchical reinforcement learning},
  author={Watanabe, Kandai and Strong, Mathew and Eldar, Omer},
  journal={arXiv preprint arXiv:2212.12786},
  year={2022}
}

@inproceedings{devlin2012dynamic,
  title={Dynamic potential-based reward shaping},
  author={Devlin, Sam Michael and Kudenko, Daniel},
  booktitle={11th International Conference on Autonomous Agents and Multiagent Systems (AAMAS 2012)},
  pages={433--440},
  year={2012},
  organization={IFAAMAS}
}

%%%%%%%%%%%%%%%%%%%%%%%%%%%%%%%%%%%%%%%%%%%%%%%%%%%%%%%%%%%%%%%%%%%%%%%%

\end{document}